# Recognition of Facial Expression Using Eigenvector Based Distributed Features and Euclidean Distance Based Decision Making Technique

Jeemoni Kalita
Department of Electronics and Communication Engineering
Assam Don Bosco University
Guwahati, India

Karen Das
Department of Electronics and Communication Engineering
Assam Don Bosco University
Guwahati, India

*Abstract*—**In this paper, an Eigenvector based system has been presented to recognize facial expressions from digital facial images. In the approach, firstly the images were acquired and cropping of five significant portions from the image was performed to extract and store the Eigenvectors specific to the expressions. The Eigenvectors for the test images were also computed, and finally the input facial image was recognized when similarity was obtained by calculating the minimum Euclidean distance between the test image and the different expressions.**

## I. INTRODUCTION

A human face carries a lot of important information while interacting to one another. In social interaction, the most common communicative hint is given by one's facial expression. Mainly in psychology, the expressions of facial features have been largely considered. As per the study of Mehrabian [1], amongst the human communication, facial expressions comprises 55% of the message transmitted in comparison to the 7% of the communication information conveyed by linguistic language and 38% by paralanguage.

This shows that the facial expression forms the major mode of interaction between the man and machine. Since for communicating the non-verbal messages the face forms the basis, the ability to read the facial emotions becomes an important part of emotional intelligence [2].

In recent years, a lot of work has been done on the affective recognition of expressions which holds the major key in the human-machine interaction. The research on the facial emotions across different cultures points out that the recognition of expressions is universal and established as constant across cultures. The first suggestion of expression of emotions as universal was given by Charles Darwin in his contriving work build from his theory of evolution. Then the psychologist Ekman and Friesen showed in their cross culture studies that the six emotions "happiness, sadness, anger, surprise, disgust and fear" are interpreted in the same way and are universal across cultures, which are known as the six basic expressions [3] [12].

In this paper, a method has been presented to design an Eigenvector based facial expression recognition system. Eigenvector based features are extracted from the images. In the training phase, a set of 10 images for each basic expression is processed and Eigenvectors specific to the expressions are stored. In the testing phase, the Eigenvector of the testing image is computed and the Euclidean distance of the Eigenvector of the testing image and all the stored Eigenvector is computed. The testing image is classified as a particular facial expression if the Euclidean distance between the Eigenvectors of that expression and the Eigenvectors of the testing image is obtained minimum compared to the Eigenvectors of the other expressions. To make the system more efficient instead of the whole image being considered, segments of the image is processed. The detail of segmentation is discussed in section IV.

## II. LITERATURE REVIEW

In 1977, Ekman and Friesen developed a famous and successful facial action coding system [4]. The Facial Action Coding System (FACS) identifies the facial muscles that cause changes in the facial expression thus enabling facial expression analysis. This system consists of 46 Action Units describing the facial behaviors. Gao, Leung, Hui, and Tananda [5] used the line based caricature of the facial expression for the line edge map (LEM) descriptor, measuring the line segment Hausdorff distance between the line caricature of the expression and the LEM of the test face. They achieved an optimal value of 86.6%, showing that the average recognition rate of females was 7.8% higher than that of males. In view of the color features, Lajevardi and Wu [6] presented a tensor based representation of the static color images. They achieved 68.8% accuracy at recognizing expression with different resolutions in CIEluv color space. A neural network is proposed in [7] that compresses the entire face region with 2-D discrete cosine transform. Ma and Khorasani [8] extended this image compression with the constructive one hidden layer neural network with the optimal block size to be 12 and the maximum number of hidden units to be 6, thus achieving the accuracy rate of nearly 93.75%.

Researchers have also used the MPEG-4 standard to provide the facial action parameters (FAPs) to represent the facial expressions. Aleksic and Katsaggelos [9] developed a facial expression recognition system utilizing these facial action parameters basically describing the eyebrow and the outer lip features, and classifying up to 93.66% of the test expressions by calculating the maximum likelihoods generated by the multistream hidden markov model (MS-HMM). Huang and He [10] presented a super resolution method to improve the face recognition of low resolution images. They applied canonical correlation analysis (CCA) to obtain the coherent features of the high resolution (HR) and low resolution (LR) images, and employed radial basis functions (RBFs) based non-linear mapping favoring the nearest neighbor (NN) classifier for recognition of single input low resolution image. The recognition rate of their method tested on the Facial Recognition Technology (FERET) face database was 84.4%, 93% for the University of Manchester Institute of Science and Technology (UMIST) database, and 95% for the Olivetti Research Laboratory (ORL) database. The approach of Eigenface method was given by Turk and Pentland [11]. Murthy and Jadon [12] enhanced this method to recognize the expression from the front view of the face, tested for the Cohn-Kanade (CK) Facial Expression database and Japanese female facial expression (JAFFE) database. Zhi, Flierl, Ruan, and Kleijn [13] applied the projected gradient method and developed the graph-preserving sparse non-negative matrix factorization (GSNMF) for extraction of feature verified on different databases. They achieved accuracy of 93.3% recognition for eyes occlusion, 94.0% for nose occlusion, 90.1% for mouth occlusion and 96.6% for images of spontaneous facial expression.

In recent years, automated recognition of facial expression has also gained popularity. Mase and Pentland [14] estimated the activity of the facial muscles using dense optical flow. In [15] this approach was extended combined with the face model, using recursive estimation and achieved an accuracy of 98%. Keith Anderson and Peter W. McOwan [16] used an enhanced ratio template algorithm to detect the frontal view of the face, and chose the multichannel gradient model (MCGM) for the motion of the face. They analyzed their recognition system using support vector machine classifier (SVM) and noted a recognition rate of 81.82%. In [17], the elastic graph matching (EGM) algorithm has been proposed and the analysis conducted for the feature extraction was a novel 2-class kernel discriminant analysis to improve the performance for the facial expression recognition. The recognition accuracy achieved for the Gabor-based elastic graph matching method was 90.5% whereas for the normalized morphological based elastic graph matching method was 91.8%. Facial expression recognition has been analyzed on visible light images, but [18] constructed a database for recognition of expression from both visible and infrared images. Gabor wavelets were also useful for recognition as it shows the enticing properties of specific spatial location and sparse object representation. Liu and Wechsler [19] presented a Gabor-Fisher based classification for face recognition using the Enhanced Fisher linear discriminant Model (EFM) along with the augmented Gabor feature, tested on 200 subjects. Zhang and Tjondronegoro [20] presented patch-based Gabor feature extraction from the automatically cropped images, in the form of patches. They matched the patches of the input image with the trained images by comparing the distance metrics and classification carried out by four different kernels SVM. The results were seen for two databases, obtaining correct recognition rate of 92.93% for JAFFE database and 94.8% for CK database. Two novel methods were proposed in [21], first detecting the dynamic facial expressions directly and second, the facial action units based detection. The classification was performed using SVMs. The recognition rate of 99.7% and 95.1% were achieved for both the methods respectively.

### III. THEORETICAL BACKGROUND

Eigenvectors and Eigenvalues are dependent on the concept of orthogonal linear transformation. An Eigenvector is basically a non-zero vector. The dominant Eigenvector of a matrix is the one corresponding to the largest Eigenvalue of that matrix. This dominant Eigenvector is important for many real world applications.

Steps used to find the features for expressions.

- Organizing the data set- Consider the data having a set of M variables that are arranged as a set of N data vectors. Thus the whole data is put into a single matrix *X* of dimensions M x N.

- Calculating the mean-

$$\mu_x = \frac{1}{N} \sum_{n=1}^{N} X[m,n] \quad (1)$$

where $\mu_x$ is the mean of the matrix *X*; m and n are indices and m=1, 2… M and n=1, 2… N

- Subtracting off the mean for each dimension-

$$X = X - \mu_x \quad (2)$$

The new matrix *X* comprises of the mean-subtracted data. The subtraction of mean is important, since it ensures that the first principal component indicates the direction of maximum variance.

- Calculating the covariance matrix-

Covariance has the same formula as that of the variance. Assume we have a 3-dimensional data set (*p, q, r*), then we can measure the covariance either between *p* and *q*, *q* and *r* or *r* and *p* dimensions. But measuring the covariance between *p* and *p*, *q* and *q*, *r* and *r* dimensions gives the value of variance of the respective *p, q, r* dimension. Variance is measured on a single dimension whereas covariance on multi-dimensions.

For 1-dimension,

$$Cov(x) = Var(x) = \frac{\sum_{i=1}^{N}(X-\mu_x)(X-\mu_x)}{N-1} \quad (3)$$

where *Var* is the variance matrix;

For 2-dimension say (x, y),

$$Cov(x,y) = \frac{\sum_{i=1}^{N}(X-\mu_x)(Y-\mu_y)}{N-1} \quad (4)$$

where *Cov(x, y)* is the covariance matrix; $\mu_y$ is the mean of another matrix Y.

- Calculating the Eigenvectors and Eigenvalues of the covariance matrix- For computing the matrix of Eigenvectors that diagonalizes the covariance matrix $C$

$$E \cdot Cov \cdot E^{-1} = D \qquad (5)$$

where $Cov$ is the covariance matrix; $E$ is the matrix of all the Eigenvectors of $Cov$, one Eigenvector per column; $D$ is the diagonal matrix of all the Eigenvalues of $Cov$ along its main diagonal, and which is zero for the rest of the elements.

The Eigenvector associated with the largest Eigenvalue displays the greatest variance in the image while the Eigenvector associated with the smallest Eigenvalue displays the least variance.

## IV. PROPOSED SYSTEM

The block diagram for the proposed system is represented in Fig. 1.

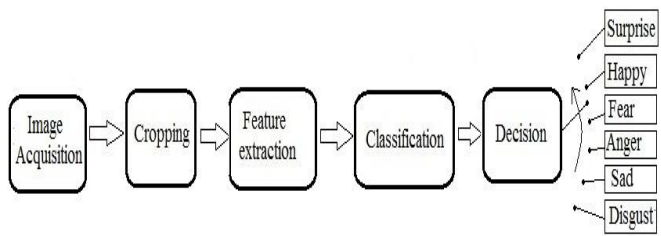

Fig. 1. Block Diagram For The Expression Recognition System

### A. Image acquisition-

Images are acquired using a digital camera. First all the images are converted into gray-scale images before going for further processing.

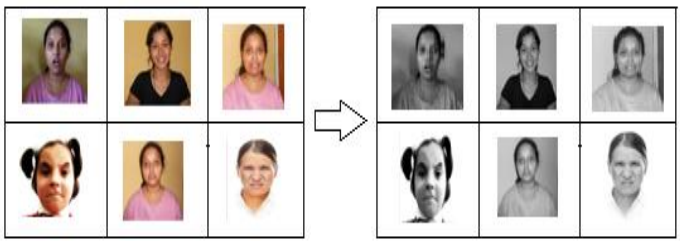

Fig. 2. Top row-Surprise, happy, fear. Bottom row-Anger, sad, disgust

### B. Cropping-

Eyes, nose and lip take different shapes for different expressions and significant information is carried by them. So instead of processing the entire face, eyes, nose and lip are processed. Before going for further processing, five significant portions are cropped from the image as shown in Fig. 3 and it shall be called as feature image.

### C. Feature extraction-

The cropped images are resized to give the value size 40 by 40 for the left and the right eye, 70 by 60 for the nose, 60 by 90 for the lip and 110 by 95 for the cropped nose and lip together. Eigenvectors are computed from these cropped images.

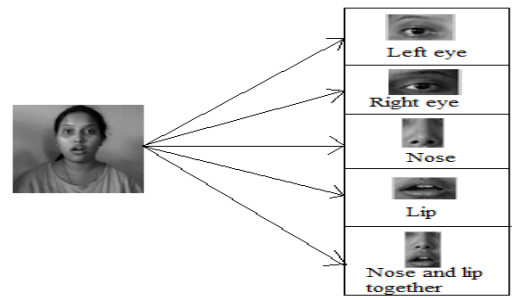

Fig. 3. Cropped images

In this work, the universal expressions are set into six classes as the training images. Eigenvectors and Eigenvalues of five different individual segments of the image is computed and stored. For a single class, after the selection of a particular feature, a matrix is obtained which is stored as, say L of dimension P x Q. Similarly for the rest of the features also, Eigenvectors and Eigenvalues are computed and stored as a matrix.

First the mean centered feature image vectors is obtained by subtracting the mean from the feature image. This image vectors are depicted as matrix only. Then the covariance matrix of each individual feature image is obtained by calculating the covariance of the matrix of each mean centered image vectors, and from each covariance matrix, the associated eigenvectors and Eigenvalues for the individually extracted features are computed.

Five significant Eigenvectors are considered for further processing which are sorted in the decreasing order of the associated Eigenvalues of the covariance matrix. With the available eigenvectors of expressions, separate subspaces for all the six universal expressions are created. With the available expression subspaces, the input image could be identified by incorporating a decision making system.

### D. Classifier-

The classifier based on the Euclidean distance has been used which is obtained by calculating the distance between the image which are to be tested and the already available images used as the training images. Then the minimum distance is observed from the set of values.

In testing, the Euclidean distance (ED) has been computed between the new (testing) image Eigenvector and the Eigen subspaces for each expression, and minimum Euclidean distance based classification is done to recognize the expression of the input image. The formula for the Euclidean distance is given by

$$ED = \sqrt{\sum(x_2 - x_1)^2} \qquad (6)$$

## V. RESULTS AND DISCUSSIONS

The testing process for the expression 'Sad' with the left eye being considered is summarized in table I. The Eigenvectors are obtained from the input image, then EDs between each Eigenvectors and the reference Eigenvectors of each trained expressions are obtained. If two expressions are same, then ED will be minimum. From minimum ED, a decision can be made of certain expression (in this case it is sad). Since five principal vectors are being considered, there will be five selections. In this case, out of the five vectors, the expression sad has been shown by two Eigenvectors and the expression fear has been shown by another two Eigenvectors.

In table II, the testing process for the expression 'Sad', with the right eye being considered is summarized. The Eigenvectors are computed from the input image and the EDs are calculated for the Eigenvectors of the input image and the Eigenvectors of the trained images. For the similar two expressions, their Euclidean distance will be minimum. Considering that, the particular expression can be decided. Since five significant vectors are taken into account, five selections are made and from table II, it is seen that the expression sad and disgust are selected most.

In table III, the testing process for the expression 'Sad', in view of the nose is summarized. The Eigenvectors of the input image are attained. The EDs of each Eigenvectors in reference to the trained expression's Eigenvectors are obtained. The ED of the same two expressions will be minimum of all the EDs. The particular expression can be determined from the minimum ED. In this case, since five principal vectors have been considered, there will be five alternatives. From table III, it has been observed that expression sad and anger has been selected the most number of times.

In table IV, the testing process for the expression 'sad', the lip being considered is summarized. From the input image, the Eigenvectors are accomplished. The EDs are estimated for each Eigenvectors in relation to the Eigenvectors for the trained images. The minimum ED is obtained for the same two expressions and from this minimum ED, the specific expression can be accomplished. In this case, since five principal vectors have been considered, there will be five selections and from table IV, it is observed that expression sad is selected thrice of all the five EDs; hence the decided expression is 'Sad'.

In table V, the testing process for the expression 'Sad' is summarized. The Eigenvectors are procured from the input image. Then EDs of each Eigenvectors are procured from the reference Eigenvectors of each trained expressions. The two expressions which are same, the value of their ED will be minimum and the preference of specific expression can be made. As five principal vectors are being considered, there will be five selections. And final decision is attained out of all the five selections. Taken into consideration the nose and the lip together, two of the Eigenvectors has exhibited the expression sad and the other two has exhibited the expression surprise.

The testing for the six basic expressions has been performed. Finally, the summarization of the values of the lowest Euclidean distance measured for the different features for the particular expression is given in table VI. The expression that gets selected more number of times is considered as the decided expression. From this table, it has been observed that the expression 'sad' has been selected the maximum number of times. Thus, a decision can be taken that the expression in the testing image is 'Sad'.

TABLE I. EUCLIDEAN DISTANCE (ED) FOR THE MASKED LEFT EYE

| Testing image | Training image | ED1 | ED2 | ED3 | ED4 | ED5 |
|---|---|---|---|---|---|---|
| 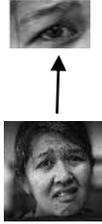 ← 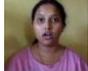 | 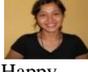 Surprise | 1.2676 | 1.8119 | 1.7794 | 1.1874 | 1.5855 |
| | 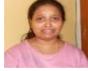 Happy | 1.4572 | 1.5182 | 1.5265 | 1.8055 | 1.6493 |
| | 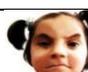 Fear | 0.8267 | **1.2712** | 1.8443 | 1.3175 | **1.2972** |
| | 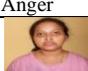 Anger | 0.9953 | 1.3782 | 1.7177 | **1.1334** | 1.4489 |
| | 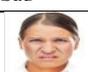 Sad | **0.7931** | 1.5523 | **1.0986** | 1.3872 | 1.4326 |
| | 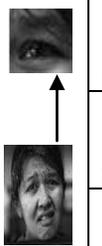 Disgust | 1.0762 | 1.6502 | 1.6997 | 1.5106 | 1.6576 |
| | Result obtained from minimum ED | Sad | Fear | Sad | Anger | Fear |

TABLE II. EUCLIDEAN DISTANCE (ED) FOR THE MASKED RIGHT EYE

| Testing image | Training image | ED1 | ED2 | ED3 | ED4 | ED5 |
|---|---|---|---|---|---|---|
| 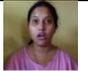 ← 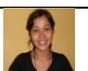 | Surprise | 1.4307 | 1.8186 | 1.1411 | 1.6590 | 1.6835 |
| | 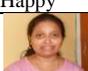 Happy | 1.6744 | 1.2974 | 1.6189 | 1.6780 | 1.3144 |
| | 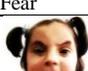 Fear | 1.1157 | 1.0956 | **0.8780** | 1.3808 | 1.5251 |
| | 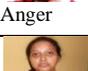 Anger | 0.9820 | 1.7105 | 1.6698 | 1.4655 | 1.4475 |
| | Sad | 1.7410 | 1.3809 | 1.6995 | **1.1275** | **1.2976** |
| | 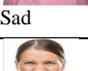 Disgust | **0.3020** | **0.8777** | 1.3580 | 1.4427 | 1.5040 |
| | Result obtained from minimum ED | Disgust | Disgust | Fear | Sad | Sad |

TABLE III. EUCLIDEAN DISTANCE (ED) FOR THE MASKED NOSE

| Testing image | Training image | ED1 | ED2 | ED3 | ED4 | ED5 |
|---|---|---|---|---|---|---|
| 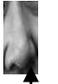 | 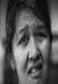 Surprise | 0.7883 | 1.7646 | 1.6823 | 1.1826 | 1.6365 |
| | 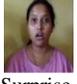 Happy | 1.8165 | 1.5430 | 1.3842 | 1.4952 | 1.5111 |
| | 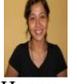 Fear | 1.7987 | 1.6020 | 1.5370 | 1.4383 | 1.6254 |
| | 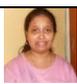 Anger | 1.0556 | 1.4425 | **1.2735** | **1.0630** | 1.6313 |
| | 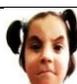 Sad | **0.7850** | 1.2584 | 1.5923 | 1.4000 | **1.2754** |
| | 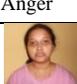 Disgust | 0.8231 | **1.2545** | 1.3076 | 1.4722 | 1.3457 |
| Result obtained from minimum ED | | Sad | Disgust | Anger | Anger | Sad |

TABLE IV. EUCLIDEAN DISTANCE (ED) FOR THE MASKED LIP

| Testing image | Training image | ED1 | ED2 | ED3 | ED4 | ED5 |
|---|---|---|---|---|---|---|
| 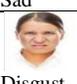 | 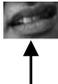 Surprise | **0.8594** | 1.7435 | 1.4553 | 1.3318 | 1.6708 |
| | 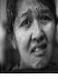 Happy | 1.6611 | 1.5389 | 1.4979 | 1.4459 | 1.4290 |
| | 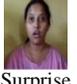 Fear | 1.6277 | 1.5100 | **1.0623** | 1.3087 | 1.6064 |
| | 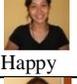 Anger | 1.2394 | 1.4300 | 1.5186 | 1.2620 | 1.3357 |
| | 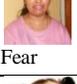 Sad | 1.1744 | **0.8795** | 1.1459 | **1.2443** | **1.2540** |
| | 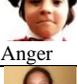 Disgust | 1.4730 | 1.5203 | 1.4403 | 1.3319 | 1.3884 |
| Result obtained from minimum ED | | Surprise | Sad | Fear | Sad | Sad |

TABLE V. EUCLIDEAN DISTANCE (ED) FOR THE MASKED NOSE AND LIP TOGETHER

| Testing image | Training image | ED1 | ED2 | ED3 | ED4 | ED5 |
|---|---|---|---|---|---|---|
| 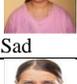 | 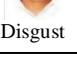 Surprise | **0.3955** | 1.9696 | 1.3062 | **1.0205** | 1.7349 |
| | 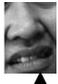 Happy | 1.8522 | 1.4260 | 1.5798 | 1.4754 | 1.5429 |
| | 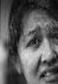 Fear | 0.8453 | 0.7654 | 1.4537 | 1.4948 | 1.4582 |
| | 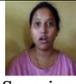 Anger | 1.0770 | 1.5184 | 1.4441 | 1.2569 | 1.5348 |
| | 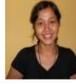 Sad | 0.7389 | 0.8601 | **1.1359** | 1.5248 | **1.1669** |
| | 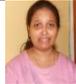 Disgust | 0.5588 | **0.6722** | 1.8764 | 1.4028 | 1.5955 |
| Result obtained from minimum ED | | Surprise | Disgust | Sad | Surprise | Sad |

In table VII, the success rate has been calculated and found to be 95% with the use of Euclidean distance based classification for 60 samples with various expressions. The time taken to process a set of image was obtained to be 0.0295 seconds.

A comparative study of our proposed work with few of the previous works performed for the recognition of facial expression has been shown in table VIII.

### VI. CONCLUSION AND FUTURE WORK

The former work on Eigen face features considered the Eigen space of the whole face. In this paper, the objective was to amend the facial expression recognition system using the Eigenvector method, creating different Eigen subspace for a distinct expression. The system has been proposed using MATLAB version 7.6.0.324 (R2008a) and Intel(R) Core(TM) i3-2330M CPU @ 2.20 GHz processor machine, Windows 7 Ultimate (32 bit), 2 GB RAM and a 14 MP camera

TABLE VI. TESTING RESULTS

| Test image | Features | Number of votes for the selected features | | | | | | Recognized expression |
|---|---|---|---|---|---|---|---|---|
| | | *Surprise* | *Happy* | *Fear* | *Anger* | *Sad* | *Disgust* | |
| 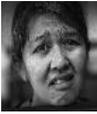 | Left eye | 0 | 0 | 2 | 1 | 2 | 0 | Sad 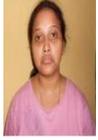 |
| | Right eye | 0 | 0 | 1 | 0 | 2 | 2 | |
| | Nose | 0 | 0 | 0 | 2 | 2 | 1 | |
| | Lip | 1 | 0 | 1 | 0 | 3 | 0 | |
| | Nose and lip together | 2 | 0 | 0 | 0 | 2 | 1 | |
| | Total votes | 3 | 0 | 4 | 3 | 11 | 4 | |

TABLE VII. SUCCESS RATE OF CLASSIFICATION

| Expression | Number of images experimented | Number of correct Recognition | Success rate |
|---|---|---|---|
| Surprise | 10 | 10 | 100% |
| Happy | 10 | 10 | 100% |
| Fear | 10 | 8 | 80% |
| Anger | 10 | 10 | 100% |
| Sad | 10 | 9 | 90% |
| Disgust | 10 | 10 | 100% |

TABLE VIII. COMPARATIVE STUDY WITH THE PREVIOUS WORKS

| Expression | Tensor perceptual color framework [6] | Eigenspaces [12] | HMM [9] | Facial movement features [20] | Our work |
|---|---|---|---|---|---|
| Anger | 62.08% | 70% | 70.6% | 87.1% | 100% |
| Disgust | 57.54% | 67% | 97.3% | 90.2% | 100% |
| Fear | 62.89% | 77% | 88.2% | 92% | 80% |
| Happy | 75.13% | 83% | 98.4% | 98.07% | 100% |
| Sad | 67.79% | 73% | 96.2% | 91.47% | 90% |
| Surprise | 84.24% | 80% | 100% | 100% | 100% |

The performance results show the efficacy of our suggested method, primarily used to recognize the six basic expressions. The recognition rate obtained for the proposed system is 95%.

As the humans can effortlessly recognize the facial expressions, the effectiveness of the machine doing the same performance without any delay is still a challenging job. Our future work is to develop a system to perform the same in real time videos.